\documentclass[letterpaper, 10 pt, journal, twoside]{IEEEtran}  % Comment this line out if you need a4paper

\IEEEoverridecommandlockouts                              % This command is only needed if 
\usepackage{graphics} % for pdf, bitmapped graphics files
\usepackage{amssymb,enumitem}
\usepackage{mathptmx} % assumes new font selection scheme installed
\usepackage{multirow}

\usepackage{times} % assumes new font selection scheme installed
\usepackage{amsmath} % assumes amsmath package installed
\usepackage{amssymb}  % assumes amsmath package installed
\usepackage{graphicx}
\usepackage{verbatim}
\usepackage[table,xcdraw]{xcolor}
\usepackage{xcolor}
\usepackage{multirow}
\usepackage{booktabs}
\usepackage{dirtytalk}

\newcommand*{\vsepfbox}[1]{%
  \begingroup
    \sbox0{\fbox{#1}}%
    \setlength{\fboxrule}{0pt}%
    \mbox{\kern-\fboxsep\fbox{\unhbox0}\kern-\fboxsep}%
  \endgroup
}

\begin{document}

\title{
% Context-Aware Error Detection: Exploring Text Prompting, activity-aware feature embedding, and Integration of Vision, Text and Kinematics
\LARGE \bf Real-Time Multimodal Activity-Aware Error Detection in Robot-Assisted Surgery
}

% \author{Anonymous Authors
% \thanks{Data and code will be released upon publication.}
% }

\author{Seyed Hamid Reza Roodabeh$^{*1}$, Zongyu Li$^{*1}$, and Homa Alemzadeh$^{1}$%
\thanks{This work has been submitted to the IEEE for possible publication. Copyright may be transferred without notice, after which this version may no longer be accessible.}
\thanks{*These authors have contributed equally.}
\thanks{$^{1}$The authors are with the Electrical and Computer Engineering Department, School of Engineering and Applied Science, University of Virginia, Charlottesville, VA, United States of America
        \{{\tt\footnotesize ydq9ag, zl7qw, ha4d\}@virginia.edu}}%
\thanks{This work was supported in part by the the National Science Foundation
(NSF) grants CNS-2146295 and CCF-2402941.}
\thanks{The source code and supplementary materials for this paper will be publicly available.}
}

% \markboth{IEEE Robotics and Automation Letters. Preprint Version. Accepted Month, Year}{}

\maketitle
% \thispagestyle{empty}
% \pagestyle{empty}

%%%%%%%%%%%%%%%%%%%%%%%%%%%%%%%%%%%%%%%%%%%%%%%%%%%%%%%%%%%%%%%%%%%%%%%%%%%%%%%%
\begin{abstract}
Robot-assisted minimally invasive surgery improves surgical precision but introduces complexity, making technical error detection essential for ensuring patient safety. Current executional error detection methods using video data often overlook fine-grained contextual descriptions of activities and error types within the hierarchical structure of surgical procedures. They also under-utilize complementary multimodal information. We propose a unified framework for executional error detection that leverages multimodal input, including video, kinematics, and descriptive textual prompts. Through activity prompting, we integrate descriptive language in gesture-level activities, instrument-object interactions, and error definitions. We also introduce activity-aware visual embeddings derived from vision encoders pretrained on surgical activity labels to compare the effectiveness of contrastive language-image embeddings with traditional image-based embeddings for error detection. By seamlessly integrating kinematic data with video and textual modalities, our framework significantly improves error detection performance. Achieving up to 5\% and 16.6\% F1 score improvements over state-of-the-art baselines on the JIGSAWS and SAR-RARP50 datasets, respectively, we demonstrate the value of combining curated textual prompts with multimodal data for accurate error detection.
\end{abstract}

%%%%%%%%%%%%%%%%%%%%%%%%%%%%%%%%%%%%%%%%%%%%%%%%%%%%%%%%%%%%%%%%%%%%%%%%%%%%%%%%
\section{INTRODUCTION}

Robot-Assisted Minimally Invasive Surgery (RAMIS) has revolutionized surgical practice by enabling complex procedures, 3D visualization, and precise control of instruments.
However, significant challenges remain in surgeon training and skill evaluation to ensure patient safety~\cite{alemzadeh2016adverse,hung2018automated}. Automated detection of technical errors, which are often due to manual execution issues even among expert surgeons~\cite{bonrath2013defining, hutchinson2022analysis}, has been an active area of research.

Existing works on automated skill assessment leverage deep learning models applied to both video and kinematic data to assess performance based on the quality and/or sequence of surgical sub-tasks and lower-level motions such as gestures~\cite{funke2019video, wang2020towards, zhang2021sd}. Despite advances in automatically predicting high-level performance metrics, such as respect for tissue, suture/needle handling, and flow of operation, these metrics alone may not sufficiently capture the technical errors that directly contribute to patient safety. 
More recent works~\cite{hutchinson2022analysis, guni2018development,xu2024sedmamba} have introduced rubrics for identifying executional and procedural errors in standard surgical tasks such as suturing. Executional errors, defined as the failure to perform a specific motor sub-task (such as gestures and motion primitives) within the procedure, are particularly shown to be challenging to detect due to subtle motion deviations and dynamics of the surgical scene~\cite{hutchinson2022analysis}. 

Early approaches to executional error detection used two-stage frameworks that segment videos into gestures before detecting errors, leveraging gesture-specific context for structured task understanding~\cite{yasar2019context, yasar2020real,li2022runtime}. This was also supported by findings that different gestures exhibit distinct error patterns~\cite{hutchinson2022analysis}. However, these approaches heavily rely on predefined gesture labels, necessitating substantial expertise and time for annotation.

Recent Vision-Language Models (VLMs) like CLIP
~\cite{clip} and Flamingo~\cite{alayrac2022flamingo} have been adapted for surgical question answering~\cite{wang2024surgical} and gesture recognition~\cite{van2022gesture}, but not explicitly for error detection. 
A recent study~\cite{wang2024surgical} leveraged textual descriptions of gestures to augment spatiotemporal video features with gesture-specific cues for error detection. However, considering the hierarchical nature of surgical procedures~\cite{hutchinson2023compass}, other critical information on finer-grained surgical activities (such as instrument-object interactions) and error types could further enhance error detection. Additionally, despite some recent efforts towards utilizing complementary semantic and contextual information from multiple modalities (e.g., video and kinematics) for activity recognition~\cite{davinciNet, weerasinghe2024multimodal, yamada2024multimodal} or skill assessment~\cite{kocielnik2023deep}, error detection methods have not yet taken full advantage of all data modalities, especially kinematics.

\begin{figure*}[t!]
    \centering
    \includegraphics[trim = 0.2in 2.4in 0.2in 2.5in, clip, width=0.9\textwidth]{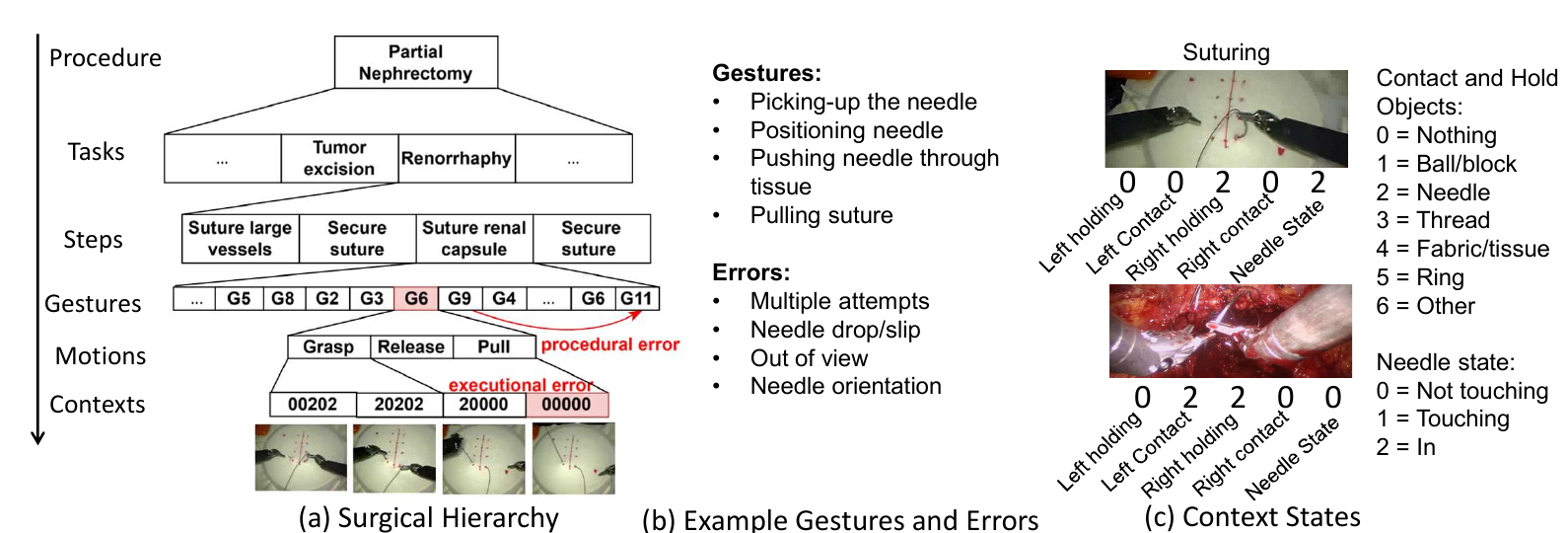}
    \vspace{-0.5em}
    \caption{Surgical hierarchy (adapted from~\cite{hutchinson2023towards}), example gesture and error labels, and context states~\cite{hutchinson2023compass}.}
    \label{fig:context-state}
    \vspace{-1em}
\end{figure*}

We propose a unified framework for real-time executional error detection that integrates video and kinematics with textual descriptions of surgical activities and errors. We introduce \textit{Activity Prompting}, which integrates gesture, instrument-object interaction, and error textual descriptions to capture multi-level surgical context during visual feature extraction. We also introduce \textit{Activity-Aware Visual Embeddings} to incorporate activity information by directly extracting features from vision encoders (e.g., ResNet~\cite{resnet}) pretrained on multi-level activity labels, and assess their effectiveness compared to contrastive language-image embeddings (e.g., using CLIP).   
Finally, we demonstrate the benefit of integrating kinematic features to further capture the fine-grained activities in error detection.

The main contributions of the paper are as follows:
\begin{itemize}
    \item We introduce activity prompting and activity-aware visual embeddings to incorporate multi-level surgical contextual information during error detection. 
    
    \item We demonstrate the effectiveness of our error detection model that integrates activity prompts including gesture definitions, associated lower-level instrument-object interactions, and error descriptions. 
    \item We show curated activity prompts integrating activity-aware textual descriptions achieve comparable or better performance than activity-aware visual embeddings generated from pretrained gesture and instrument-object interaction models as feature extractors.
   
    \item We demonstrate that our model outperforms the baselines with up to 5\% and 16.6\% improvements in F1 score on the JIGSAWS~\cite{gao2014jhu} and SAR-RARP50~\cite{xu2024sedmamba} datasets, respectively.
\end{itemize}

\section{BACKGROUND and RELATED WORK}

\textbf{Surgical procedures} can be decomposed into steps, tasks, gestures, and motions~\cite{hutchinson2023compass}, as shown in Fig.~\ref{fig:context-state} (a). Gestures are semantically meaningful activity segments tied to a specific intent. Fig.~\ref{fig:context-state} (b) shows representative suturing gestures, along with common executional error types. To represent lower-level surgical activity,~\cite{hutchinson2023compass} defines the surgical context in dry-lab tasks using five state variables. The first four describe the object held by or in contact with each grasper: left holding (\(S_1\)), left contact (\(S_2\)), right holding (\(S_3\)), and right contact (\(S_4\)). These variables encode interactions with objects such as nothing (0), ball/block (1), needle (2), thread (3), fabric/tissue (4), ring (5), or other scene objects (6), depending on the task. The fifth variable (\(S_5\)) is task specific and describes progress within a trial, such as the needle state in Fig.~\ref{fig:context-state} (c), where 0, 1, and 2 denote not touching, touching, and in, respectively. With these definitions, each gesture can be represented as a sequence of atomic motions that induce context-state changes, or equivalently as a lower-level sequence of context states. Fig.~\ref{fig:context-state} (a) illustrates this for the gesture G6, pulling suture with the left hand. Since executional errors occur at finer levels of the surgical hierarchy, they can appear as anomalies in motion sequences~\cite{hutchinson2023towardsMP} or in context-state transitions. This motivates incorporating rich sub-gesture representations into error detection. To incorporate sub-gesture context into error detection, we convert the surgical context-state labels~\cite{hutchinson2023compass}, shown in Fig.~\ref{fig:context-state} (c), into concise natural-language descriptions of instrument-object interactions, such as which grasper is holding or contacting the needle, thread, or tissue. These interaction descriptions are used to generate activity-aware textual prompts for the proposed detection framework and to define interaction labels for pretraining the activity-aware visual encoders.

\textbf{Error Detection} from surgical videos has recently gained attention in RAMIS. Hutchinson et al.~\cite{hutchinson2022analysis} introduced rubrics and labeled the publicly available JIGSAWS dataset~\cite{gao2014jhu} for procedural and executional errors. Since procedural errors can often be detected from deviations in predefined gesture transitions or grammar graphs, subsequent work has mainly focused on executional errors. Yasar et al.~\cite{yasar2020real} and Li et al.~\cite{li2022runtime} proposed task- and gesture-specific error detection models using kinematic data, while Shao et al.~\cite{shao2024think} introduced Chain-of-Gesture (CoG) prompting to integrate gesture-level textual descriptions with video. More recently, Xu et al.~\cite{xu2024sedmamba} proposed SEDMamba, a vision-based model using state-space modeling with fine-to-coarse temporal fusion, and provided error annotations for an in-vivo RAMIS dataset. Although these works show the importance of activity information, they do not explore surgical activity representations below the gesture level.
% ~\cite{hutchinson2023compass,nwoye2023cholectriplet2021, neumuth2011modeling}

Recent studies emphasize the hierarchical structure of surgical procedures and the value of lower-level representations, including motion primitives~\cite{hutchinson2023compass,hutchinson2023towardsMP}, action triplets~\cite{nwoye2023cholectriplet2021}, and surgical context states~\cite{hutchinson2023compass,hutchinson2023towards}, for RAMIS tasks~\cite{hutchinson2023evaluating}. In activity recognition and skill assessment, detecting fine-grained motions and tool-object interactions has been shown to provide richer surgical scene understanding~\cite{nwoye2023cholectriplet2021,hutchinson2023towardsMP, hutchinson2023evaluating}. However, current error detection pipelines remain largely gesture-level and do not incorporate sub-gesture hierarchical information. Another limitation is online applicability: SEDMamba processes full videos offline, while CoG, despite its low inference latency of approximately 7 ms, uses non-causal components that preclude live video-stream processing. Our framework addresses this gap by targeting online inference with a practical trade-off between detection granularity, precision, and real-time efficiency.

\textbf{Vision-Language Models} have also been adopted in surgical applications, including Surgical-LVLM~\cite{wang2024surgical}, which uses specialized visual perception low-rank adaptation for domain-specific question answering, and PeskaVLP~\cite{srivastavprocedure}, which introduces procedure-encoded surgical knowledge-augmented video-language pretraining for surgical gesture recognition. However, these models do not directly address executional error detection. The Gesture-Visual Reasoning (GVR) module~\cite{shao2024think} integrates gesture-level information with video for error detection using contrastive VLMs and transformer attention~\cite{vaswani2017attention}, but its prompts are limited to gesture descriptions and omit lower-level motion and error descriptions. More recently,~\cite{cares} introduced a collaborative agentic framework for error detection, showing that standalone VLMs perform relatively poorly compared to supervised alternatives, but can become competitive when augmented with surgical domain knowledge.
%yadav2021review

\textbf{Multimodal Data Fusion} using vision, audio, and motion data has been widely studied in human activity recognition~\cite{sun2022human}. In RAMIS, multimodal deep learning has been applied to activity recognition, trajectory prediction, and feedback classification~\cite{van2022gesture, davinciNet, weerasinghe2024multimodal, yamada2024multimodal, kocielnik2023deep}. These studies show that kinematic data can complement video by providing precise measurements of fine-grained instrument motion and interactions. Nevertheless, there remains a gap in combining complementary fine-grained activity information from video, kinematics, and text for surgical error detection.

\begin{figure*}[t!]
    \centering
    % \vspace{0.75in}
    %left, bottom, right, top
    \includegraphics[width=\textwidth]{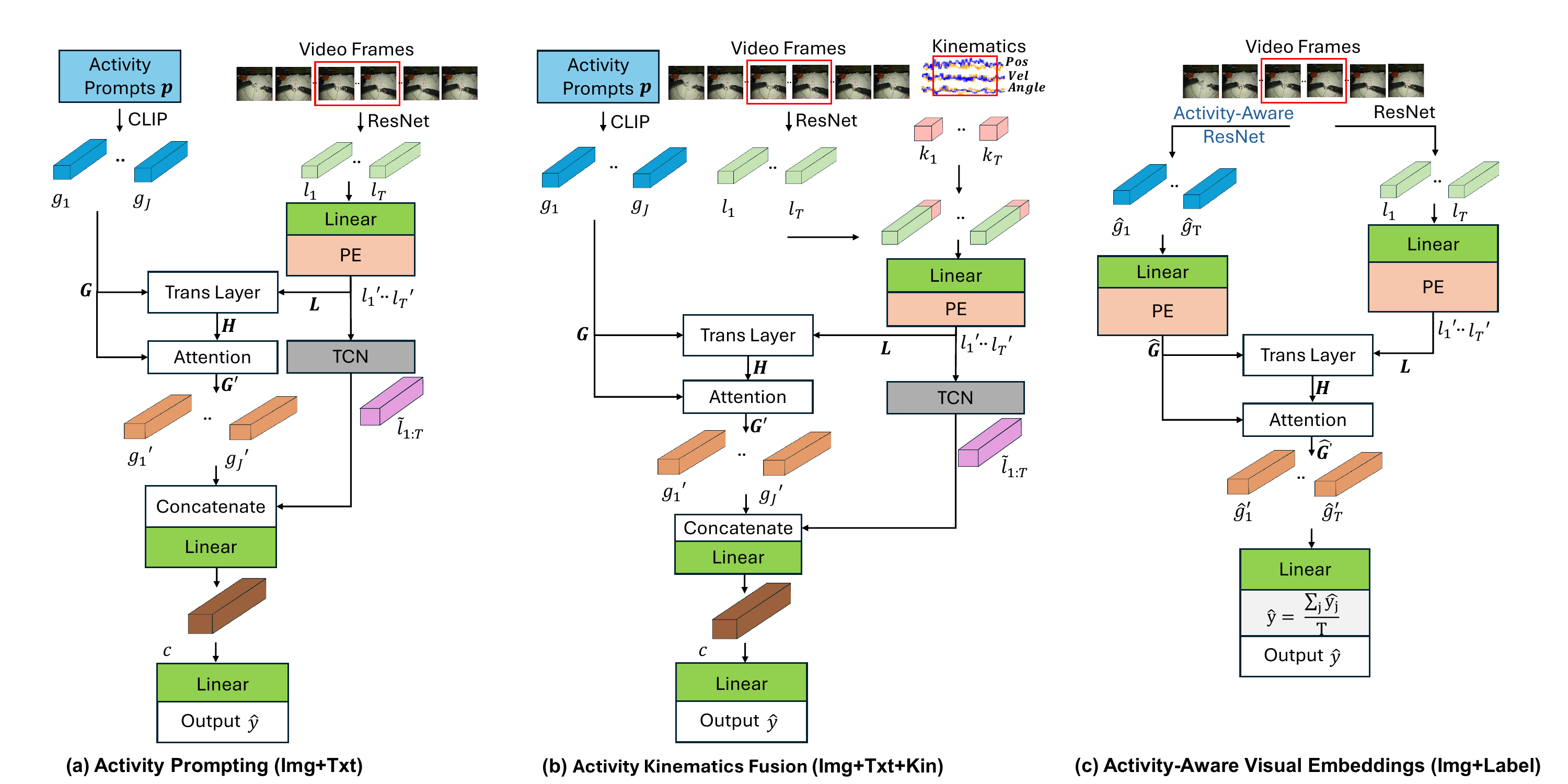}
    \vspace{-1em}
    \caption{Activity-Aware Error Detection Pipeline. Abbreviations: \textit{Pos: position, Vel: velocity, Angle: grasper angle, Img: image data, Txt: textual prompts, Kin: kinematic data, Label: gesture or interaction labels}.}
    \label{fig:pipeline}
    \vspace{-1em}
\end{figure*}

\section{Methods}
\subsection{Problem Formulation}
% Our goal is to detect surgical errors in RAMIS by leveraging multimodal data, including video, kinematics, and textual prompts. We frame the problem as a binary classification task, where given an input window of multimodal time-series data of length $T$, including Red-Green-Blue (RGB) video frames \(\{f_j\}_{j=1}^T\) (capturing the visual context of the surgical task) and kinematic data \(\{k_j\}_{j=1}^T\) (representing robot motions such as position and velocity of surgical instruments), the aim is to predict in real-time whether an error occurred (\(y = 1\)) or not (\(y = 0\)). 

% Our goal is to detect surgical errors in RAMIS by leveraging multimodal data, including video, kinematics, and textual prompts. We frame the problem as a binary classification task, where given an input window of multimodal time-series data of length $T$, including Red-Green-Blue (RGB) video frames \(\{f_j\}_{j=1}^T\) and kinematic data \(\{k_j\}_{j=1}^T\), we aim to detect in real-time whether an error has occurred (\(y = 1\)) or not (\(y = 0\)).

Our goal is to detect surgical errors in RAMIS by leveraging multimodal data, including video, kinematics, and textual prompts. We frame the problem as a binary classification task, where given an input window of multimodal time-series data of length $T$, including Red-Green-Blue (RGB) video frames and robot kinematics, we aim to detect in real-time whether an error has occurred (\(y = 1\)) or not (\(y = 0\)).

\subsection{Activity-Aware Multimodal Fusion}
\label{sec:activity-fusion}
We propose a unified deep-learning framework based on pretrained vision encoders, contrastive vision-language encoders, and transformers with attention layers to fuse video and kinematic data with multi-level surgical activity information for error detection. Specifically, we encode the information on surgical gestures, instrument-object interactions, and error types as textual prompts or as activity labels using pretrained vision encoders. These \textit{activity embeddings} are then used to contextualize the spatial embeddings from video or fused video/kinematic data for activity-aware error detection. 

We study three different setups for incorporating the activity information as shown in Figure~\ref{fig:pipeline}: 

\begin{enumerate}[label=(\alph*)]
\item Activity Prompting (\textit{Img+Txt})
\item Activity Kinematics Fusion (\textit{Img+Txt+Kin})
\item Activity-Aware Visual Embeddings (\textit{Img+Label})
\end{enumerate}

In setups (a) and (b), we adopt a similar architecture as the Gesture-Visual Reasoning (GVR) module in~\cite{shao2024think}, with added modifications for better input feature injection and enhanced model expressiveness. In the setup (c), we substitute encoded textual prompts by a fine-tuned vision encoder path, which generates per-frame activity-aware embeddings. 
The major components of the framework include the following:

\subsubsection{Visual Feature Extraction} 
In each setup, a sequence of video frames \(\mathbf{F} =\{f_j\}_{j=1}^T \in \mathbb{R}^{T \times 3 \times H \times W}\), where \(T\) is the number of frames, 3 is the number of RGB channels, and \(H\) and \(W\) are the height and width of each frame, are passed through a ResNet50 backbone \cite{resnet} pretrained on ImageNet. ResNet50 is chosen as it achieves the best balance between inference latency and predictive accuracy (see Appendix C), but similar vision encoders may also be used. The extracted spatial visual embeddings are represented as: 
{\setlength{\abovedisplayskip}{4pt}
\setlength{\belowdisplayskip}{4pt}
\[
\ell_{j} = \text{ResNet}(f_j), \quad \ell_{j} \in \mathbb{R}^{d_{model}}
\]
}
where  \(\mathbf{\ell}_j\) represents the embedding for the \(j\)-th frame, and \(d_{model}\) is the dimension of the extracted embeddings.

To capture the sequential nature of video frames, positional encoding is added to the obtained visual embeddings:
{\setlength{\abovedisplayskip}{4pt}
\setlength{\belowdisplayskip}{4pt}
\[
\mathbf{\ell}_j^{'} = \mathbf{\ell}_j + \text{PE}(j), \quad j = 1, \dots, T 
\]
}
where \(\text{PE}(j) \in \mathbb{R}^{d_\text{model}}\) represents the positional encoding vector.
% where the positional encoding, \(\text{PE}(j) \in \mathbb{R}^{d_\text{model}}\), is defined as:
% {\setlength{\abovedisplayskip}{4pt}
% \setlength{\belowdisplayskip}{4pt}
% \[
% \begin{aligned}
% \text{PE}(j, 2i)   &= \sin\left(\frac{j}{10000^{2i/d_{\mathrm{model}}}}\right), \\
% \text{PE}(j, 2i+1) &= \cos\left(\frac{j}{10000^{2i/d_{\mathrm{model}}}}\right),
% \end{aligned}
% \qquad 0 \le i < \frac{d_{\mathrm{model}}}{2}
% \]
% }

%  \subsubsection{Vision and Kinematics Fusion}
% In the \textit{Img+Txt+Kin} setup (Figure \ref{fig:pipeline} (b)), the kinematic features, denoted as \(\{{k}_j\}_{j=1}^T\), and spatial video embeddings \(\{\ell_j\}_{j=1}^T\) are first transformed via a linear projection followed by a non-linear ReLU activation $\sigma$ to map them into the same \(d_\text{model}\) dimensional space. They are then concatenated and further transformed by an additional linear layer to create a unified spatial feature vector \(\ell'_{j} \in \mathbb{R}^{d_\text{model}}\). To capture the sequential nature of video frames, positional encoding is added directly to this unified vector:
% \[
% \ell'_{j} = W_u [\sigma(W_v \ell_{j} + b_v) ; \sigma(W_k k_{j} + b_k)] + b_u + \text{PE}(j)
% \]
\subsubsection{Vision and Kinematics Fusion}
In the \textit{Img+Txt+Kin} setup, we combine spatial video embeddings \(\{\ell_j\}_{j=1}^T\) with kinematic features \(\{k_j\}_{j=1}^T\). Each kinematic vector \(k_j \in \mathbb{R}^{14}\) includes the Cartesian position \((x,y,z)\), velocity \((v_x,v_y,v_z)\), and grasper angle for both surgical instruments. The video and kinematic features are first projected onto the same \(d_\text{model}\)-dimensional space using linear layers followed by a ReLU activation \(\sigma\). The resulting features are then concatenated and passed through an additional linear layer to obtain a unified spatial feature vector \(\ell'_j \in \mathbb{R}^{d_\text{model}}\). Positional encoding is added directly to this unified representation:
{\setlength{\abovedisplayskip}{4pt}
\setlength{\belowdisplayskip}{4pt}
\[
\ell'_{j} = W_u [\sigma(W_v \ell_{j} + b_v) ; \sigma(W_k k_{j} + b_k)] + b_u + \text{PE}(j)
\]
}

\subsubsection{Activity Contextualization}
\label{sec:activity_contextualization}

We contextualize the spatial feature sequence with multi-level activity information using a cross-attention module within a transformer encoder layer. Let \(
\mathbf{L} = [\mathbf{\ell}_1^{'}; \dots; \mathbf{\ell}_T^{'}] \in \mathbb{R}^{T \times d_{\mathrm{model}}}
\)
denote the sequence of spatial embeddings. For the \textit{Img+Txt} and \textit{Img+Txt+Kin} setups, we obtain a set of activity-prompt embeddings
\[
\mathbf{G} = [\mathbf{g}_1; \dots; \mathbf{g}_J] \in \mathbb{R}^{J \times d_{\mathrm{model}}}
\]
where \(J\) is the number of prompts and each \(\mathbf{g}_j\) is generated by the CLIP text encoder~\cite{clip}. For the \textit{Img+Label} setup, we instead use activity-aware visual embeddings
\[
\hat{\mathbf{G}} = [\hat{\mathbf{g}}_1; \dots; \hat{\mathbf{g}}_T] \in \mathbb{R}^{T \times d_{\mathrm{model}}}
\]
where each \(\hat{\mathbf{g}}_j\) is extracted from an activity-aware ResNet50 model, pretrained on activity labels (gesture or instrument-object interaction labels), thus embedding activity-level information directly into the visual features:
\[
\hat{\mathbf{g}}_j = \mathrm{ResNet}_{\mathrm{activity}}(f_j), \quad \hat{\mathbf{g}}_j \in \mathbb{R}^{d_{\mathrm{model}}}
\]
For the prompt-based setups (Figure \ref{fig:pipeline} (a) and (b)), the first attention operation is implemented using a transformer encoder layer, where the activity embeddings (\(\mathbf{G}\)) serve as queries and the spatial embeddings (\(\mathbf{L}\))  serve as keys and values:
{\setlength{\abovedisplayskip}{4pt}
\setlength{\belowdisplayskip}{4pt}
\[
\mathbf{H} = \mathrm{Attention}(\mathbf{G}, \mathbf{L}, \mathbf{L})
= \mathrm{softmax}\!\left(\frac{\mathbf{G}\mathbf{L}^{\top}}{\sqrt{d_{\mathrm{model}}}}\right)\mathbf{L}
\]
}
where \(\mathbf{H} = [\mathbf{h}_1; \dots; \mathbf{h}_J] \in \mathbb{R}^{J \times d_{\mathrm{model}}}\) contains the activity-contextualized prompt representations. We then refine the original prompt embeddings through a second attention operation:
\vspace{-6pt} 
\[
\mathbf{G}^{'} = \mathrm{Attention}(\mathbf{H}, \mathbf{G}, \mathbf{G})
\]
where \(\mathbf{G}^{'} = [\mathbf{g}_1^{'}; \dots; \mathbf{g}_J^{'}] \in \mathbb{R}^{J \times d_{\mathrm{model}}}\)

To inject dynamic temporal information from the input sequence, we further compute a pooled spatial feature using a standard Temporal Convolutional Network (TCN) \cite{tcn} encoder:
\[
\tilde{\mathbf{\ell}}_{1:T} = \mathrm{TCN}(\mathbf{L}) \in \mathbb{R}^{d_{\mathrm{model}}}
\]
which summarizes the spatial input over the full window. This pooled feature is concatenated with the refined prompt embeddings and projected back to \(d_{\mathrm{model}}\):
\[
\mathbf{c} = \sigma\!\left( W_f
\begin{bmatrix}
\mathbf{g}_1^{'}; \dots; \mathbf{g}_J^{'}; \tilde{\mathbf{\ell}}_{1:T}
\end{bmatrix}
+ \mathbf{b}_f \right)
\]
This fusion step provides a direct path from the input sequence to the final representation, preventing the model from relying only on a fixed combination of prompt embeddings and enabling the prediction to remain conditioned on the current visual and kinematic context.

Finally, the fused representation is passed through a linear layer with sigmoid activation to produce the binary error prediction:
\[
\hat{y} = \sigma(\mathbf{W}\mathbf{c} + \mathbf{b})
\]
where \(\hat{y}\) denotes the probability of error occurrence for the current input window.

% \textcolor{blue}{
% For the \textit{Img+Label} setup (Figure \ref{fig:pipeline} (c)), the same transformer-encoder attention mechanism is used, but with the activity-aware visual embeddings as the query sequence:%
% \[
% \mathbf{H}_{\mathrm{activity}} = \mathrm{Attention}(\hat{\mathbf{G}}, \mathbf{L}, \mathbf{L}),
% \]
% where \(\mathbf{H}_{\mathrm{activity}} = [\mathbf{h}^{\mathrm{act}}_1; \dots; \mathbf{h}^{\mathrm{act}}_T] \in \mathbb{R}^{T \times d_{\mathrm{model}}}\). Since each activity-aware embedding is aligned with its corresponding frame, the model directly produces a frame-level prediction from each contextualized embedding:%
% \[
% \hat{y}_j = \sigma(\mathbf{W}\mathbf{h}^{\mathrm{act}}_j + \mathbf{b}), \quad j = 1, \dots, T.
% \]
% Thus, \(\hat{y}_j\) represents the probability of error occurrence for frame \(j\) within the input window. The window-level prediction is obtained by averaging the frame-level predictions:%
% \[
% \hat{y} = \frac{1}{T} \sum_{j=1}^{T} \hat{y}_{j},
% \]}

For the \textit{Img+Label} setup (Figure \ref{fig:pipeline} (c)), the same transformer-encoder attention mechanism is used, except activity-aware visual embeddings \(\hat{\mathbf{G}}^{'} = [\hat{\mathbf{g}}_1^{'}; \dots; \hat{\mathbf{g}}_T^{'}] \in \mathbb{R}^{T \times d_{\mathrm{model}}}\) have replaced the activity-prompt embeddings:%
\[
\hat{\mathbf{G}}^{'} = \mathrm{Attention}(\mathbf{H}, \mathbf{\hat G}, \mathbf{\hat G})
\]
% where
% {\setlength{\abovedisplayskip}{4pt}
% \setlength{\belowdisplayskip}{4pt}
% \[
% \hat{\mathbf{G}}^{'} = [\hat{\mathbf{g}}_1^{'}; \dots; \hat{\mathbf{g}}_T^{'}] \in \mathbb{R}^{T \times d_{\mathrm{model}}}
% \]
% }
Since each activity-aware embedding is aligned with its corresponding frame, the model directly produces a frame-level prediction from each contextualized embedding:%
\[
\hat{y}_j = \sigma(\mathbf{W}\hat{\mathbf{g}}_j^{'} + \mathbf{b}), \quad j = 1, \dots, T
\]
Thus, \(\hat{y}_j\) represents the probability of error occurrence for frame \(j\) within the input window. The window-level prediction is obtained by averaging the frame-level predictions:%
\vspace{-6pt} 
\[
\hat{y} = \frac{1}{T} \sum_{j=1}^{T} \hat{y}_{j}
\]

% \textcolor{blue}{
% To train the model, we use a class-weighted binary cross-entropy loss:%
\vspace{-6pt} 
% \[
% \mathcal{L}_{\mathrm{WBCE}} = -\frac{1}{N} \sum_{n=1}^{N}
% \left[
% w_{1} y_{n} \log(\hat{y}_{n}) +
% w_{0} (1-y_{n}) \log(1-\hat{y}_{n})
% \right],
% \]
% where \(y_{n} \in \{0,1\}\) is the ground-truth label, \(\hat{y}_{n} \in [0,1]\) is the predicted probability, $N$ is number of training samples, and \(w_{1}\) and \(w_{0}\) are the weights for the positive and negative classes, respectively, and are obtained from the class frequencies of the train set.
% }
To train the model, we use a class-weighted binary cross-entropy loss, where the weights for the positive and negative classes are obtained from the class frequencies within the training set.
% \textcolor{blue}{
% To train the model, we use a class-weighted binary cross-entropy loss:%
% % \[
% % \mathcal{L}_{\mathrm{WBCE}} = -\frac{1}{N} \sum_{n=1}^{N}
% % \left[
% % w_{1} y^{(n)} \log(\hat{y}^{(n)}) +\\
% % w_{0} (1-y^{(n)}) \log(1-\hat{y}^{(n)})
% % \right],
% % \]
% {\setlength{\abovedisplayskip}{4pt}
% \setlength{\belowdisplayskip}{4pt}
% \[
% \begin{aligned}
% \mathcal{L}_{\mathrm{WBCE}} = -\frac{1}{N} \sum_{n=1}^{N} \Big[
% &\, w_{1} y^{(n)} \log(\hat{y}^{(n)}) \\
% &+ w_{0} (1-y^{(n)}) \log(1-\hat{y}^{(n)})
% \Big]
% \end{aligned}
% \]
% }
% where \(y^{(n)} \in \{0,1\}\) is the ground-truth label and \(\hat{y}^{(n)} \in [0,1]\) is the predicted probability for the \(n\)-th training sample, \(N\) is the number of training samples, and \(w_{1}\) and \(w_{0}\) are the weights for the positive and negative classes, respectively, obtained from the class frequencies of the training set.
% }

\subsection{Multi-level Activity and Error Prompts}
\label{sec:prompts}
For activity contextualization using prompts, we encode the multi-level activity information as textual prompts, including gesture prompts, similar to those used in~\cite{shao2024think}, and descriptions of instrument-object interactions and error types: 
\begin{itemize}
    \item \textbf{Gesture Prompts:} These prompts describe specific surgical gestures: \emph{``a surgeon is [gesture]'' }. %(e.g. ``A surgeon is positioning the needle tip")
    
    \item \textbf{Instrument-Object Interaction Prompts:} These prompts describe specific low-level interactions between surgical instruments and objects: \emph{``a surgeon is [interacting with the object] with [instrument]''}. See Appendix A for more details. %(e.g. ``A surgeon is holding the needle with the right grasper")
    
    \item \textbf{Error Prompts:} These prompts explicitly focus on surgical errors : \emph{``a surgeon is [error]'' }. %(e.g. ``A surgeon is dropping the needle")
    
    \item \textbf{Gesture + Error Prompts:} These prompts include gesture and gesture-specific error descriptions:  \emph{``a surgeon is [gesture] but [error]''}. %(e.g.,  ``A surgeon is positioning the needle tip but makes multiple attempts.")
    
    \item \textbf{Interaction + Error Prompts}: These prompts provide low-level interaction descriptions, formulated based on context state changes for each gesture along with associated error information. Each prompt is in the form: “\emph{a surgeon is [interacting with the object] to [goal] with [instrument] but [error]}”. %(e.g. ``A surgeon is holding the needle with the left grasper to let the right grasper contact the thread but the needle drops.")
\end{itemize}

The activity embeddings  \(\{g_j^{}\}_{j=1}^J\) can be obtained from textual prompts  \(\{p_j^{}\}_{j=1}^J\) with a CLIP-based text encoder:%
\[
g_j = \text{CLIP}(p_{j}), \quad \mathbf{g}_j \in \mathbb{R}^{d_\text{model}}.
\]
The full set of prompts for all prompt types is available in the supplementary materials, Appendix B.

\section{EXPERIMENTS}

\subsection{Datasets and Evaluation Metrics}
We evaluate our methods using a dry-lab dataset (JIGSAWS~\cite{gao2014jhu}) and a real surgical dataset (SAR-RARP50~\cite{xu2024sedmamba}). More details about the size and label distribution of each dataset is presented in Appendix D.

\subsubsection{JIGSAWS} We focus on 67 trials of suturing and needle passing tasks in JIGSAWS. These tasks include annotated executional error labels from~\cite{hutchinson2022analysis}. Additionally, we leverage synchronized kinematic and video data to train the error detection models, and frame-level manual annotations of gestures from~\cite{gao2014jhu} and context states from~\cite{hutchinson2023compass} to fine-tune the activity-aware ResNet vision encoders. 
%Gesture labels from~\cite{gao2014jhu} and instrument-object interaction labels derived from context state labels from~\cite{hutchinson2023compass} are used for training the pretrained ResNet to obtain activity-aware visual embeddings. 
% The gestures, errors, and instrument-object interactions derived from context state definitions are used to generate gesture, instrument-object interaction and error prompts.
For Gesture + Error prompts, the gesture-specific error table from~\cite{hutchinson2022analysis} is used to generate prompts that simultaneously include both gesture descriptions and their corresponding errors for the JIGSAWS dataset. %The complete set of gestures and multi-level prompts for JIGSAWS are available in Appendix B.

\subsubsection{SAR-RARP50} We focus on 48 videos of the suturing task with annotated error and gesture labels from the SAR-RARP50 dataset. We create the context state labels based on the definitions in~\cite{hutchinson2023compass}, using an adjusted context detection method based on~\cite{li2023robotic}. The context detection method first utilizes the segmentation masks for the graspers, the needle, and the thread provided by the SAR-RARP50 segmentation challenge~\cite{psychogyios2023sar}. Then, logic rules similar to those in~\cite{li2023robotic} are used to determine the value held by each context-state variable.
For constructing Gesture + Error prompts, we use ChatGPT to generate gesture-specific errors for the SAR-RARP50 dataset, as it does not provide a predefined set of gesture-specific errors. Given the SAR-RARP50 gesture definitions, we ask ChatGPT what are the relevant errors for each of the gestures according to errors provided in~\cite{xu2024sedmamba}.
Since the SAR-RARP50 dataset focuses on the suturing task and shares similar gestures with the JIGSAWS dataset, as in Figure \ref{fig:context-state} (b), 
we use similar descriptions for the context states and instrument-object interactions and combine them with the generated gesture-specific errors to create the Interaction + Error prompts. 
% Similar to JIGSAWS, the gesture and instrument-object interaction labels, and derived context state labels for the SAR-RARP50 dataset are used for training ResNet to obtain activity-aware visual embeddings.

\subsubsection{Metrics} We use Leave-One-Supertrial-Out (LOSO) cross-validation~\cite{gao2014jhu} for the JIGSAWS dataset, where the \(i\)-th trial of each surgeon is excluded from the dataset to serve as the test set. For the SAR-RARP50 dataset, 40 videos are used as training set and 8 videos are used as test set~\cite{xu2024sedmamba}, and we repeat each run 10 times with different random seeds to obtain the mean and standard deviation of each metric. To evaluate the performance of error detection, we report the binary F1 score, accuracy, and Jaccard index for both datasets. Results for the Leave-One-User-Out (LOUO) cross validation~\cite{gao2014jhu} are presented in Appendix E.

\subsection{Experimental Setup}
\vspace{-0.4em}
We conducted our experiments on a single NVIDIA RTX 3090 Ti with 24 GB of GPU memory. The video and kinematics data were preprocessed by downsampling to 5 Hz and resizing the video frames to 240×240 pixels, followed by center cropping to 224×224 pixels. Input images were normalized using standard ImageNet statistics (mean=[0.485, 0.456, 0.406], std=[0.229, 0.224, 0.225]) to align with the pre-processing requirements of ResNet models~\cite{resnet}. The kinematics data were standardized using mean and standard deviation normalization for each variable. The input sequence length and stride were set to $T=10$ samples (2-second windows) and 6 samples (1.2-second strides), respectively. A window is labeled erroneous if it contains at least one erroneous frame. This window is short enough to enable real-time, granular error detection, yet long enough to provide the model with sufficient temporal context.

The ResNet models used to generate activity-aware visual embeddings were fine-tuned for per-frame activity recognition. The same training data as the down-stream error detection model (the training portion of each JIGSAWS LOSO split, and the training set of the SAR-RARP50 dataset) was used for training ResNet models to predict the underlying activity (gesture or instrument-object interaction) of that frame. The supervised fine-tuning of the models was performed for 10 epochs, using a learning rate of $10^{-5}$ and the AdamW optimization algorithm~\cite{loshchilov2017decoupled}.

For training, input windows from different trials and tasks were shuffled together for all three error detection models to support more balanced learning. All models were trained with a batch size of 64 using the AdamW optimizer and a learning rate of $5 \times 10^{-4}$. Models using only image and text inputs were trained for 50 epochs, while models incorporating kinematic data were trained for 100 epochs. Throughout training, all encoders were kept frozen, including the ResNet backbones used for spatial and activity-aware feature extraction and the CLIP text encoder used to generate prompt embeddings. Only the remaining components of the model were updated through gradient-based optimization.
% To train all three error detection models, input windows were mixed across trials and tasks to promote balanced learning. \textcolor{blue}{A batch size of 64 and the AdamW optimization algorithm with a learning rate of $5 \times 10^{-4}$ was used.} The model was trained for 50 epochs when using only text and image inputs, and for 100 epochs when kinematic data were also included. During training, the encoders used in each setup, including the ResNet vision encoders for spatial and activity-aware feature extraction and the CLIP text encoder for prompt generation, are kept frozen. Only the remaining model components are updated through gradient-based optimization. %This design allows the framework to integrate activity prompts or activity-aware visual embeddings while reducing the risk of overfitting.

\subsection{Experimental Results} 
The results in Tables \ref{tab:jigsaws} and \ref{tab:sarrarp} highlight the effectiveness of different setups for integrating activity information with video data and their impact on error detection performance. 

\subsubsection{Observations on Activity Prompting} Among the prompt configurations evaluated on JIGSAWS, the Gesture+Error prompts achieve the best performance, with an F1 score of 0.745, outperforming activity-only Gesture prompts (F1: 0.712), Interaction prompts (F1: 0.724), Error-only prompts (F1: 0.733), and the ResNet-only baseline (F1: 0.719). A Pearson correlation analysis showed no significant relationship between the number of prompts and F1 score, indicating that prompt quality and semantic content are more important than prompt quantity. Adding error-specific information to the activity-only prompt sets consistently improves detection performance. The use of Interaction prompts also shows that richer descriptions of instrument-object relationships provide more useful context for error detection than basic gesture descriptions alone. On SAR-RARP50, the ResNet-only baseline obtains an F1 score of 0.590, while activity prompting improves performance up to 0.746 with Interaction+Error prompts, corresponding to a 16.6\% increase over SOTA and surpassing both Interaction-only and Error-only prompts.

\begin{table}[t!]
\vspace{0.15cm}
\centering\small
\renewcommand{\arraystretch}{1.4} % Increases vertical height of the rows
\caption{Error detection performance on JIGSAWS.}
\resizebox{\columnwidth}{!}{%
\begin{tabular}{|l|l|l|l|c|c|c|}
\hline
\textbf{Setup} & \textbf{Activity Knowledge} & \textbf{Inputs} & \textbf{Encoders} & \textbf{F1} & \textbf{Accuracy} & \textbf{Jaccard} \\
\hline
Baseline (ResNet)  & -- & Img & ResNet & 0.719\,\tiny$\pm$0.005 & 0.658\,\tiny$\pm$0.011 & 0.591\,\tiny$\pm$0.013 \\
\hline
\hline
\multirow{5}{*}{\shortstack{Activity \\ Prompting}}
& Gesture & \multirow{5}{*}{Img+Txt} & \multirow{5}{*}{\shortstack{CLIP+\\ResNet}} & 0.712\,\tiny$\pm$0.056 & 0.655\,\tiny$\pm$0.040 & 0.555\,\tiny$\pm$0.072 \\
\cline{2-2}\cline{5-7}
& Interaction & & & 0.724\,\tiny$\pm$0.052 & 0.664\,\tiny$\pm$0.031 & 0.570\,\tiny$\pm$0.066 \\
\cline{2-2}\cline{5-7}
& Error & & & 0.733\,\tiny$\pm$0.051 & 0.675\,\tiny$\pm$0.033 & 0.598\,\tiny$\pm$0.065 \\
\cline{2-2}\cline{5-7}
& Gesture + Error & & & \textbf{0.745\,\tiny$\pm$0.047} & \textbf{0.692\,\tiny$\pm$0.032} & \textbf{0.588\,\tiny$\pm$0.059} \\
\cline{2-2}\cline{5-7}
& Interaction + Error & & & \textbf{0.730\,\tiny$\pm$0.055} & \textbf{0.671\,\tiny$\pm$0.035} & \textbf{0.553\,\tiny$\pm$0.068} \\
\hline
\hline
\multirow{2}{*}{\shortstack{Activity-Aware \\ Embeddings}}
& Interaction pretrained & \multirow{2}{*}{Img+Label} & \multirow{2}{*}{ResNet x 2} & 0.710\,\tiny$\pm$0.055 & 0.671\,\tiny$\pm$0.035 & 0.553\,\tiny$\pm$0.068 \\
\cline{2-2}\cline{5-7}
& Gesture pretrained & & & 0.695\,\tiny$\pm$0.058 & 0.591\,\tiny$\pm$0.056 & 0.535\,\tiny$\pm$0.069 \\
\hline
\hline
\multirow{5}{*}{\shortstack{Activity \\ Kinematics \\ Fusion}}
& Gesture & & & 0.729\,\tiny$\pm$0.047 & 0.666\,\tiny$\pm$0.040 & 0.565\,\tiny$\pm$0.060 \\
\cline{2-2}\cline{5-7}
& Interaction & \multirow{5}{*}{\shortstack{Img+Txt\\+Kin}} & \multirow{5}{*}{\shortstack{CLIP+\\ResNet}} & 0.724\,\tiny$\pm$0.054 & 0.687\,\tiny$\pm$0.025 & 0.570\,\tiny$\pm$0.068 \\
\cline{2-2}\cline{5-7}
& Error & & & 0.745\,\tiny$\pm$0.048 & 0.702\,\tiny$\pm$0.062 & 0.591\,\tiny$\pm$0.061 \\
\cline{2-2}\cline{5-7}

& Gesture + Error & & & \textbf{0.760\,\tiny$\pm$0.044} & \textbf{0.717\,\tiny$\pm$0.027} & \textbf{0.634\,\tiny$\pm$0.056} \\
\cline{2-2}\cline{5-7}
& Interaction + Error & & & \textbf{0.749\,\tiny$\pm$0.043} & \textbf{0.699\,\tiny$\pm$0.027} & \textbf{0.606\,\tiny$\pm$0.054} \\
\hline
\end{tabular}}
\label{tab:jigsaws}
\vspace{-1.5em}
\end{table}

\begin{table}[tb]
\vspace{0.15cm}
\centering\small
\renewcommand{\arraystretch}{1.4} % Increases vertical height of the rows
\caption{Error detection performance on SAR-RARP50.}
\resizebox{\columnwidth}{!}{%
\begin{tabular}{|l|l|l|l|c|c|c|}
\hline
\textbf{Setup} & \textbf{Activity Knowledge} & \textbf{Inputs} & \textbf{Encoders} & \textbf{F1} & \textbf{Accuracy} & \textbf{Jaccard} \\
\hline
Baseline (ResNet) & -- & Img & ResNet & 0.590\,\tiny$\pm$0.031 & 0.470\,\tiny$\pm$0.024 & 0.490\,\tiny$\pm$0.045 \\
\hline
\hline
\multirow{5}{*}{\shortstack{Activity \\ Prompting}}
& Gesture & \multirow{5}{*}{Img+Txt} & \multirow{5}{*}{\shortstack{CLIP+\\ResNet}} & 0.735\,\tiny$\pm$0.033 & 0.612\,\tiny$\pm$0.042 & 0.593\,\tiny$\pm$0.051 \\
\cline{2-2}\cline{5-7}
& Interaction & & & 0.713\,\tiny$\pm$0.052 & 0.563\,\tiny$\pm$0.041 & 0.554\,\tiny$\pm$0.060 \\
\cline{2-2}\cline{5-7}
& Error & & & 0.707\,\tiny$\pm$0.048 & 0.565\,\tiny$\pm$0.038 & 0.547\,\tiny$\pm$0.047 \\
\cline{2-2}\cline{5-7}
& Gesture + Error & & & 0.712\,\tiny$\pm$0.028 & 0.578\,\tiny$\pm$0.036 & 0.553\,\tiny$\pm$0.014 \\
\cline{2-2}\cline{5-7}
& Interaction + Error & & & \textbf{0.746\,\tiny$\pm$0.021} & \textbf{0.623\,\tiny$\pm$0.017} & \textbf{0.596\,\tiny$\pm$0.039} \\
\hline
\hline
\multirow{2}{*}{\shortstack{Activity-Aware \\ Embeddings}}
& Gesture Embedding & \multirow{2}{*}{Img+Label} & \multirow{2}{*}{ResNet x 2} & 0.709\,\tiny$\pm$0.035 & 0.557\,\tiny$\pm$0.044 & 0.549\,\tiny$\pm$0.052 \\
\cline{2-2}\cline{5-7}
& Interaction Embedding & & & 0.717\,\tiny$\pm$0.029 & 0.578\,\tiny$\pm$0.031 & 0.559\,\tiny$\pm$0.040 \\
\hline
\end{tabular}}
\label{tab:sarrarp}
\vspace{-2.3em}
\end{table}

\begin{table*}[tb]
\vspace{0.15cm}
\centering
\caption{Comparison with state-of-the-art methods on JIGSAWS and SAR-RARP50.}
\resizebox{0.95\textwidth}{!}{%
\begin{tabular}{|l|l|l|l|l|l|l|c|}
\hline
\textbf{Method} &\textbf{ Activity Knowledge} & \textbf{Inputs} & \textbf{Dataset} &\textbf{F1} & \textbf{Accuracy} & \textbf{Jaccard} & \textbf{\% Improvement (F1)} \\ \hline
Siamese-LSTM (G*T*)~\cite{li2022runtime} & - & Kin & \multirow{7}{*}{JIGSAWS} & 0.700 ± 0.010 & 0.650 ± 0.010 & 0.530 ± 0.020 & -3.3 \\ \cline{1-1} \cline{2-2} \cline{3-3} \cline{5-8}
ResNet & - & \multirow{2}{*}{Img} & & 0.719 ± 0.005 & 0.658 ± 0.011 & 0.591 ± 0.013 & -0.7 \\ \cline{1-2} \cline{5-8}
SEDMamba~\cite{xu2024sedmamba} & - & & & 0.683 ± 0.048 & 0.634 ± 0.041 & 0.522 ± 0.073 & -5.7 \\ \cline{1-1} \cline{2-2} \cline{3-3} \cline{5-8}
CoG (GVR only)~\cite{shao2024think} & Gestures & \multirow{3}{*}{Img+Txt} & & 0.724 ± 0.049 & 0.663 ± 0.047 & 0.570 ± 0.057 & 0 \\ \cline{1-1} \cline{2-2} \cline{5-8}
\multirow{3}{*}{Ours} & Gesture+Error & & & 0.745 ± 0.047 & 0.692 ± 0.032 & 0.588 ± 0.059 & 2.9 \\ \cline{2-2} \cline{5-8}
& Interaction+Error & & & 0.730 ± 0.055 & 0.671 ± 0.035 & 0.553 ± 0.068 & 0.8 \\ \cline{2-2}\cline{3-3} \cline{5-8}
& Gesture+Error & Img+Txt+Kin & & \textbf{0.760 ± 0.044} & \textbf{0.717 ± 0.027} & \textbf{0.634 ± 0.056} & \textbf{5.0} \\ \hline \hline
ResNet & - & \multirow{2}{*}{Img} & \multirow{5}{*}{SAR-RARP50} & 0.590 ± 0.031 & 0.470 ± 0.024 & 0.490 ± 0.045 & -7.8 \\ \cline{1-1} \cline{2-2} \cline{5-8}
SEDMamba~\cite{xu2024sedmamba} & - & & & 0.640 ± 0.015 & \textbf{0.610 ± 0.022} & 0.370 ± 0.038 & 0 \\ \cline{1-1} \cline{2-2} \cline{3-3} \cline{5-8}
CoG (GVR only)~\cite{shao2024think} & Gestures & \multirow{3}{*}{Img+Txt} & & 0.580 ± 0.041 & 0.680 ± 0.019 & 0.410 ± 0.027 & -9.4 \\ \cline{1-1} \cline{2-2} \cline{5-8}
\multirow{2}{*}{Ours} & Gesture+Error & & & 0.712 ± 0.028 & 0.578 ± 0.036 & 0.553 ± 0.014 & 11.3 \\ \cline{2-2} \cline{5-8}
& Interaction + Error & & & \textbf{0.746 ± 0.021} & \textbf{0.623 ± 0.017} & \textbf{0.596 ± 0.039} & \textbf{16.6} \\ \hline
\end{tabular}%
}
\label{tab:sota}
\vspace{-2em}
\end{table*}

These results underscore the benefit of more complex prompts that blend fine-grained behavioral intents with error conditions. Combining detailed, low-level interaction descriptions and error-related contextual information strengthens the model's ability to isolate errors.
%These results underscore how more complex prompts, blending intricate behavioral intents with fault conditions, are beneficial. Combining gesture information, especially detailed, lower-level descriptions of gestures, and error-related contextual information fortifies the model's ability to isolate errors.

\noindent\vsepfbox{
    \parbox{0.95\linewidth}{
    \textbf{Insight 1:} Prompt specificity and relevance are critical for extracting meaningful contextual features.}  
    }

\subsubsection{Observations on Activity Kinematics Fusion} Because SAR-RARP50 does not provide kinematic data, kinematics fusion is evaluated only on JIGSAWS. The clearest gain appears when kinematics are combined with Gesture+Error prompts, raising the F1 score from 0.745 to 0.760, which is the best overall result. This suggests that adding kinematic information helps anchor abstract textual representations in measurable physical motion, supplying complementary cues that may not be fully captured by visual features alone.
% We show the kinematics fusion results on the JIGSAWS dataset because SAR-RARP50 does not contain kinematics data. The most notable improvement is observed when integrating kinematics with Gesture+Error prompts, resulting in the highest overall F1 score of 0.760, compared to 0.745 without kinematics. This highlights how kinematic fusion helps to ground abstract text representations with concrete physical measurements, providing complementary motion cues that visual features may lack.
\noindent\vsepfbox{
    \parbox{0.95\linewidth}{
    \textbf{Insight 2:} Incorporating multi-level activity information, including gestures and their corresponding instrument–object interactions and lower level motion kinematics, combined with error prompts, provides a rich contextual framework for surgical error detection.}  
    }

\subsubsection{Observations on Activity-Aware Visual Embeddings} On JIGSAWS, the Activity-Aware Gesture and Interaction embedding variants achieve F1 scores of 0.695 and 0.710, respectively, which are about 1–2\% lower than their corresponding Activity Prompting models based on textual prompts. On SAR-RARP50, the Activity-Aware Visual Embedding variants for Gestures and Interactions reach F1 scores of 0.709 and 0.717, giving performance that is similar to, or slightly below, the corresponding prompt-based setups. These results suggest that strong error detection performance does not necessarily depend on costly and time-intensive activity annotation. Instead, textual prompts grounded in domain knowledge about surgical activities, such as gestures and interactions, and enriched with medically informed error associations, can match or outperform pretrained models and two-stage training pipelines.
% On JIGSAWS, Activity-Aware Gesture (F1: 0.710) and Interaction (F1: 0.695) embedding methods underperform the corresponding Activity Prompting setups that use textual prompts by \%1-2. On SAR-RARP50, Activity-Aware Visual Embeddings of Gestures and Interactions achieve F1 scores of 0.709 and 0.717, respectively, with near or slightly lower scores compared to the corresponding activity prompts. This shows that accurate error detection does not necessarily require expensive and time consuming activity annotation, and textual prompts, informed by domain knowledge of activities such as gestures and interactions, and the extension of such knowledge with midically-informed and relevant error associations can achieve performance that is superior to pre-trained models and two-stage training pipelines.

% Using curated textual prompts with a CLIP text encoder yields comparable or better performance than Activity-Aware Visual Embeddings, especially with error-focused prompts. On JIGSAWS, Error prompts (F1:0.733), Gesture+Error prompts (F1: 0.745), and Interaction+Error prompts (F1: 0.730) show strong performance. On SAR-RARP50, low-level Interaction+Error prompts achieve the best result (F1: 0.746). This establishes that better curated language prompts are superior to activity-aware embeddings, especially considering the latter rigidly relies on massive upfront activity annotations for pretraining.
\noindent\vsepfbox{
    \parbox{0.95\linewidth}{
    \textbf{Insight 3:} Contrastive language-image embeddings based on fine-grained activity prompts can achieve similar or better error detection performance to visual embeddings, while simplifying training and inference by eliminating the need for activity labeling and pretraining.}  
    }

\subsection{State-of-the-Art Comparison} 

Table~\ref{tab:sota} compares our proposed multimodal activity-aware error detection framework with state-of-the-art (SOTA) surgical error detection methods, including the kinematics-only Siamese-LSTM (G*T*)~\cite{li2022runtime}, image-only ResNet~\cite{resnet} trained for error detection, CoG~\cite{shao2024think}, and SEDMamba~\cite{xu2024sedmamba}, on the JIGSAWS and SAR-RARP50 datasets. For JIGSAWS, CoG is used as the primary SOTA baseline, while for SAR-RARP50, SEDMamba serves as the main baseline. The SOTA results shown are obtained using the original authors' implementations.

On JIGSAWS, our activity-aware error detection framework with different prompt sets outperforms all baselines and the SOTA. When kinematic data are incorporated together with video inputs, the model using the Gesture+Error prompting strategy achieves the highest overall performance in our experiments, with an F1 score of 0.760. This corresponds to a \(5\%\) improvement over the baseline, despite using a context window of only 2 seconds, which is \(25\%\) of the temporal context used by the CoG architecture. We also outperform the entire CoG pipeline (GVR + multi-scale temporal reasoning) by around 1.5\% in F1 score.

\begin{figure}[t!]
    \centering
    \includegraphics[width=\columnwidth]{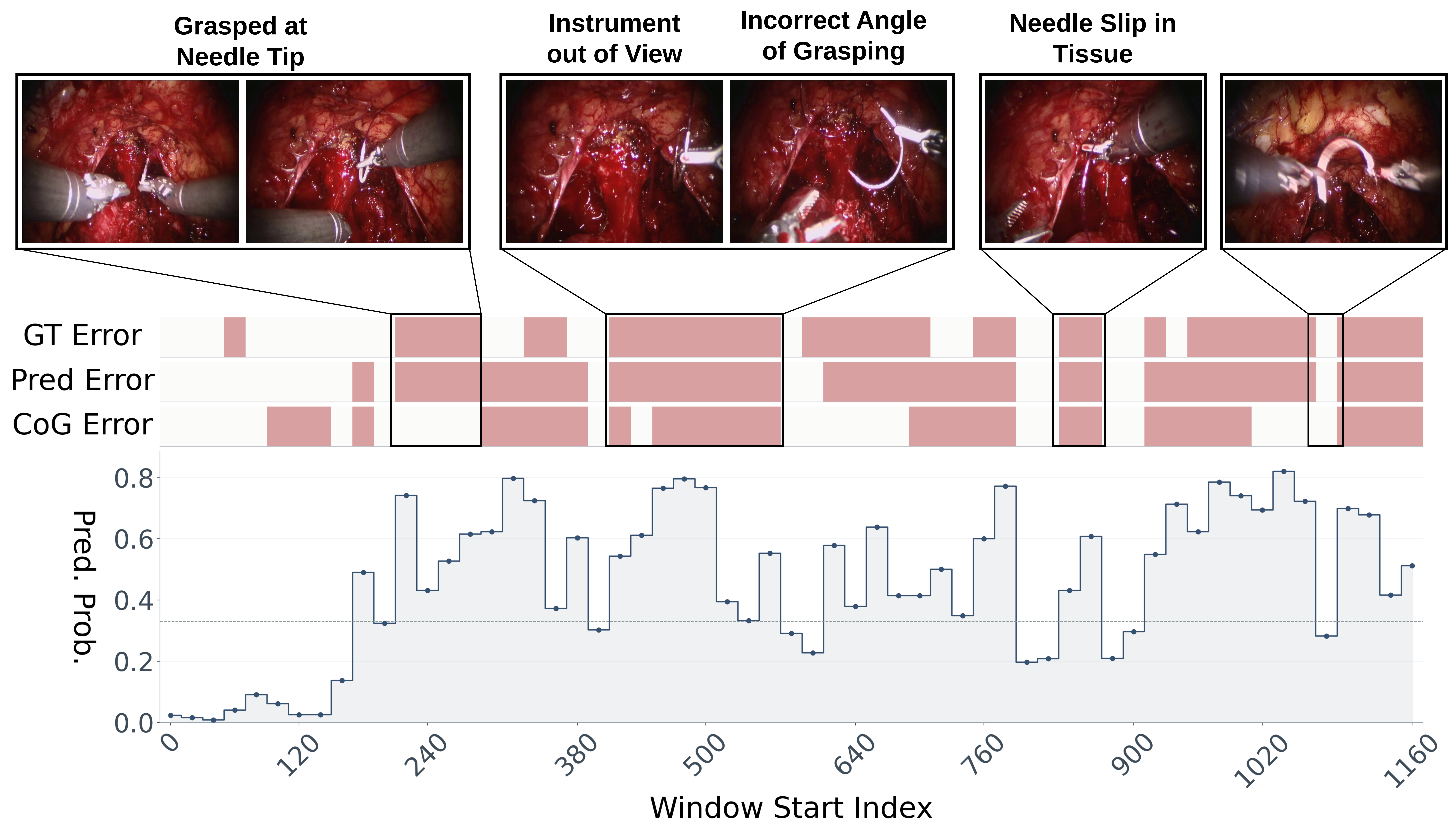}
    \caption{Example error-detection output of our framework on a test video from the SAR-RARP50 dataset. Ground-truth error segments and predicted error segments by our model and the CoG model are shown in red on the \textit{GT Error}, \textit{Pred Error} and \textit{CoG Error} bars, respectively, with our model’s prediction probability plotted below.}
    \label{fig:error_time_series}
    \vspace{-1.5em}
\end{figure}

Similarly, on SAR-RARP50, the Interaction+Error prompting strategy achieves the best F1 performance, improving over the SOTA baseline by \(16.6\%\). Compared with the ResNet baseline, SEDMamba~\cite{xu2024sedmamba} improves F1 score and accuracy but yields a lower Jaccard index. CoG obtains the highest accuracy among all models; however, accuracy can be misleading for imbalanced datasets such as SAR-RARP50, which contains approximately \(30\%\) error frames and \(70\%\) normal frames, or approximately \(40\%\) erroneous and \(60\%\) normal samples after windowing. Therefore, more balanced metrics such as F1 score are more informative for this setting, and our models achieve the strongest performance under this criterion. As a demonstrative example, Fig. \ref{fig:error_time_series} shows the error predictions and output of our model against the ground-truth, and the predictions of the CoG model. Both models miss the first ground-truth error: the probability rises immediately after the error appears but remains insufficiently confident. All later error instances are at least partially detected, with the first confident detection near window 220 for \say{Grasped at Needle Tip}, despite a constrained, narrow needle view; CoG appears to miss this event entirely. The model also confidently detects \say{Instrument out of View} and \say{Incorrect angle of grasping}, and pinpoints a difficult \say{Needle slip in tissue} in a cluttered, bloody scene. Near the video end, both our model and CoG correctly classify a true negative despite motion-induced blur and low instrument-tip/needle contrast.

% \begin{table}[h]
% \vspace{0.15cm}
% \centering\small
% \caption{Maximum Computational complexity and inference latency per 10-sample window.}
% \resizebox{\columnwidth}{!}{%
% \color{blue}
% \begin{tabular}{|l|c|c|c|}
% \hline
% \textbf{Model} & \textbf{Params (M)} & \textbf{FLOPs (M)} & \textbf{Latency (ms)} \\ \hline
% Activity-Aware Visual Embeddings & 4.2 & 84 & 1.2 \\ \hline
% Activity Prompting & 9.7 & 245 & 1.6 \\ \hline
% Activity Kinematics Fusion & 10.0 & 257 & 1.7 \\ \hline
% \end{tabular}%
% }
% \label{tab:latency}
% \vspace{-1.5em}
% \end{table}

\begin{table}[tb]
\vspace{0.15cm}
\centering\small
\renewcommand{\arraystretch}{1.4}
\caption{Model complexity and latency comparison. Enc: vision encoder, FLOPS: Floating Point Operations Per Second, Lat: latency}
\resizebox{\columnwidth}{!}{%
\begin{tabular}{|l|l|c|c|c|c|c|}
\hline
\textbf{Setup} & \begin{tabular}[b]{@{}l@{}}\textbf{Activity} \\ \textbf{Knowledge}\end{tabular} & \begin{tabular}[b]{@{}c@{}}\textbf{Enc.} \\ \textbf{Params}\end{tabular} & \begin{tabular}[b]{@{}c@{}}\textbf{Model} \\ \textbf{Params}\end{tabular} & \textbf{FLOPS} & \begin{tabular}[b]{@{}c@{}}\textbf{Model} \\ \textbf{Lat. (ms)}\end{tabular} & \begin{tabular}[b]{@{}c@{}}\textbf{E2E} \\ \textbf{Lat. (ms)}\end{tabular} \\
\hline
CoG & Gesture & 23.5M & 1.8M & 81.8B & 17.06\,\tiny$\pm$0.02 & 51.18\,\tiny$\pm$0.58 \\
\hline
SEDMamba & -- & 1.1B & 290.0K & 5.9T & 0.7\,\tiny$\pm$0.03 & 368.06\,\tiny$\pm$0.52 \\
\hline
\hline
Activity Prompting & Interaction + Error & 23.5M & 9.7M & 82.0B & 1.53\,\tiny$\pm$0.01 & 35.72\,\tiny$\pm$0.59 \\
\hline
\shortstack{\textbf{Act. Kin. Fusion}} & \textbf{Gesture + Error} & \textbf{23.5M} & \textbf{10.0M} & \textbf{82.0B} & \textbf{1.63}\,\tiny$\pm$0.01 & \textbf{35.55}\,\tiny$\pm$0.03 \\
\hline
\shortstack{Act.-Aware Embeddings} & -- & 47M & 42.0M & 82.6B & 1.36\,\tiny$\pm$0.00 & 41.59\,\tiny$\pm$0.68 \\
\hline
\end{tabular}}
\label{tab:latency}
\vspace{-2em}
\end{table}

Furthermore, we evaluated the computational complexity and per-window end-to-end inference latency of our models compared to the SOTA to assert their real-time applicability, as summarized in Table \ref{tab:latency}. This latency includes input loading and preprocessing, encoder latency, and model inference latency. It is observed that our best performing model, Activity Kinematics Fusion with Gesture+Error prompts, has the lowest end-to-end latency of around 36 ms, allowing near-30 Hz execution the of error detection pipeline. SEDMamba has the smallest and fastest model, but uses a computationally expensive image encoder (DinoV2-Giant), and while CoG's per frame execution is around 2 ms, its end-to-end latency for a window of 10 frames is larger than all of our models. A more comprehensive complexity profile of all model configurations is presented in Appendix F.

\section{CONCLUSION}

This work presents a multimodal framework for real-time surgical error detection that integrates fine-grained, hierarchical activity descriptions with video and kinematic data. By explicitly modeling gestures, lower-level instrument-object interactions, and error semantics, our approach achieves SOTA performance, underscoring the critical role of multimodal context in characterizing complex surgical scenes.

% A primary insight from our study is the effectiveness of \textit{activity prompting} compared to \textit{activity-aware visual embeddings}. Learning visual embeddings requires dense frame-level annotations and a computationally expensive pretraining stage for the vision encoder. In contrast, our prompting strategy completely eliminates this annotation and compute overhead while maintaining, and often exceeding, predictive accuracy. We further demonstrate that compound textual prompts, those jointly integrating low-level activity descriptions with explicit error conditions, provide the most discriminative context for the model. We also observed that fusing raw kinematic data yields moderate gains by supplying concrete motion cues that complement visual data.

A key finding is that \textit{activity prompting} can match or outperform \textit{activity-aware visual embeddings} without dense frame-level activity labels or separate encoder pretraining. Prompts that combine activity context with explicit error descriptions provide the strongest discriminative cues, while kinematic fusion offers additional, moderate gains by adding instrument-motion information that complements visual data.

Despite these advancements, our approach has several limitations. First, achieving peak performance with kinematic fusion restricts the framework's broader applicability, given the limited availability of surgical datasets containing synchronized video and robot kinematics. New multimodal RAMIS datasets may help address this limitation once error annotations become available \cite{midas, imitatecholec}. Second, the success of our activity prompting heavily relies on the manual curation and quality of the textual prompts. Finally, while we validate on suturing and needle-passing tasks, generalization to less structured and more diverse procedures remains to be shown.

\bibliographystyle{IEEEtran}
\bibliography{bib}

@article{shao2024think,
  title={Think step by step: Chain-of-gesture prompting for error detection in robotic surgical videos},
  author={Shao, Zhimin and Xu, Jialang and Stoyanov, Danail and Mazomenos, Evangelos B and Jin, Yueming},
  journal={IEEE Robotics and Automation Letters},
  year={2024},
  publisher={IEEE}
}

@inproceedings{gao2014jhu,
  title={Jhu-isi gesture and skill assessment working set (jigsaws): A surgical activity dataset for human motion modeling},
  author={Gao, Yixin and Vedula, S Swaroop and Reiley, Carol E and Ahmidi, Narges and Varadarajan, Balakrishnan and Lin, Henry C and Tao, Lingling and Zappella, Luca and B{\'e}jar, Benjam{\i}n and Yuh, David D and others},
  booktitle={MICCAI workshop: M2cai},
  volume={3},
  number={2014},
  pages={3},
  year={2014}
}

@article{hutchinson2023compass,
  title={COMPASS: a formal framework and aggregate dataset for generalized surgical procedure modeling},
  author={Hutchinson, Kay and Reyes, Ian and Li, Zongyu and Alemzadeh, Homa},
  journal={International Journal of Computer Assisted Radiology and Surgery},
  volume={18},
  number={12},
  pages={2143--2154},
  year={2023},
  publisher={Springer}
}

@article{hutchinson2022analysis,
  title={Analysis of executional and procedural errors in dry-lab robotic surgery experiments},
  author={Hutchinson, Kay and Li, Zongyu and Cantrell, Leigh A and Schenkman, Noah S and Alemzadeh, Homa},
  journal={The International Journal of Medical Robotics and Computer Assisted Surgery},
  volume={18},
  number={3},
  pages={e2375},
  year={2022},
  publisher={Wiley Online Library}
}

@article{hutchinson2023evaluating,
  title={Evaluating the task generalization of temporal convolutional networks for surgical gesture and motion recognition using kinematic data},
  author={Hutchinson, Kay and Reyes, Ian and Li, Zongyu and Alemzadeh, Homa},
  journal={IEEE Robotics and Automation Letters},
  volume={8},
  number={8},
  pages={5132--5139},
  year={2023},
  publisher={IEEE}
}

@article{hutchinson2023towardsMP,
  title={Towards Interpretable Motion-level Skill Assessment in Robotic Surgery},
  author={Hutchinson, Kay and Chen, Katherina and Alemzadeh, Homa},
  journal={arXiv preprint arXiv:2311.05838},
  year={2023}
}

@inproceedings{weerasinghe2024multimodal,
  title={Multimodal transformers for real-time surgical activity prediction},
  author={Weerasinghe, Keshara and Roodabeh, Seyed Hamid Reza and Hutchinson, Kay and Alemzadeh, Homa},
  booktitle={2024 IEEE International Conference on Robotics and Automation (ICRA)},
  pages={13323--13330},
  year={2024},
  organization={IEEE}
}

@article{davinciNet,
  title={daVinciNet: Joint Prediction of Motion and Surgical State in Robot-Assisted Surgery},
  author={Qin, Yidan and Feyzabadi, Seyedshams Feyzabadi and Allan, Max and Burdick, Joel W. and Azizian, Mahdi },
  year={2020},
  publisher={IEEE}
}

@article{yamada2024multimodal,
  title={Multimodal semi-supervised learning for online recognition of multi-granularity surgical workflows},
  author={Yamada, Yutaro and Colan, Jacinto and Davila, Ana and Hasegawa, Yasuhisa},
  journal={International Journal of Computer Assisted Radiology and Surgery},
  volume={19},
  number={6},
  pages={1075--1083},
  year={2024},
  publisher={Springer}
}

@article{van2022gesture,
  title={Gesture recognition in robotic surgery with multimodal attention},
  author={Van Amsterdam, Beatrice and Funke, Isabel and Edwards, Eddie and Speidel, Stefanie and Collins, Justin and Sridhar, Ashwin and Kelly, John and Clarkson, Matthew J and Stoyanov, Danail},
  journal={IEEE transactions on medical imaging},
  volume={41},
  number={7},
  pages={1677--1687},
  year={2022},
  publisher={IEEE}
}

@inproceedings{kocielnik2023deep,
  title={Deep multimodal fusion for surgical feedback classification},
  author={Kocielnik, Rafal and Wong, Elyssa Y and Chu, Timothy N and Lin, Lydia and Huang, De-An and Wang, Jiayun and Anandkumar, Anima and Hung, Andrew J},
  booktitle={Machine Learning for Health (ML4H)},
  pages={256--267},
  year={2023},
  organization={PMLR}
}

@article{sun2022human,
  title={Human action recognition from various data modalities: A review},
  author={Sun, Zehua and Ke, Qiuhong and Rahmani, Hossein and Bennamoun, Mohammed and Wang, Gang and Liu, Jun},
  journal={IEEE transactions on pattern analysis and machine intelligence},
  volume={45},
  number={3},
  pages={3200--3225},
  year={2022},
  publisher={IEEE}
}

@article{alemzadeh2016adverse,
  title={Adverse events in robotic surgery: a retrospective study of 14 years of FDA data},
  author={Alemzadeh, Homa and Raman, Jaishankar and Leveson, Nancy and Kalbarczyk, Zbigniew and Iyer, Ravishankar K},
  journal={PloS one},
  volume={11},
  number={4},
  pages={e0151470},
  year={2016},
  publisher={Public Library of Science San Francisco, CA USA}
}

@article{hung2018automated,
  title={Automated performance metrics and machine learning algorithms to measure surgeon performance and anticipate clinical outcomes in robotic surgery},
  author={Hung, Andrew J and Chen, Jian and Gill, Inderbir S},
  journal={JAMA surgery},
  volume={153},
  number={8},
  pages={770--771},
  year={2018},
  publisher={American Medical Association}
}

@article{guni2018development,
  title={Development of a technical checklist for the assessment of suturing in robotic surgery},
  author={Guni, Ahmad and Raison, Nicholas and Challacombe, Ben and Khan, Shamim and Dasgupta, Prokar and Ahmed, Kamran},
  journal={Surgical endoscopy},
  volume={32},
  number={11},
  pages={4402--4407},
  year={2018},
  publisher={Springer}
}

@article{xu2024sedmamba,
  title={Sedmamba: Enhancing selective state space modelling with bottleneck mechanism and fine-to-coarse temporal fusion for efficient error detection in robot-assisted surgery},
  author={Xu, Jialang and Sirajudeen, Nazir and Boal, Matthew and Francis, Nader and Stoyanov, Danail and Mazomenos, Evangelos B},
  journal={IEEE Robotics and Automation Letters},
  year={2024},
  publisher={IEEE}
}

@inproceedings{hutchinson2023towards,
  title={Towards surgical context inference and translation to gestures},
  author={Hutchinson, Kay and Li, Zongyu and Reyes, Ian and Alemzadeh, Homa},
  booktitle={2023 IEEE International Conference on Robotics and Automation (ICRA)},
  pages={6802--6809},
  year={2023},
  organization={IEEE}
}

@inproceedings{yasar2019context,
  title={Context-aware monitoring in robotic surgery},
  author={Yasar, Mohammad Samin and Evans, David and Alemzadeh, Homa},
  booktitle={2019 International symposium on medical robotics (ISMR)},
  pages={1--7},
  year={2019},
  organization={IEEE}
}

@inproceedings{li2022runtime,
  title={Runtime detection of executional errors in robot-assisted surgery},
  author={Li, Zongyu and Hutchinson, Kay and Alemzadeh, Homa},
  booktitle={2022 International conference on robotics and automation (ICRA)},
  pages={3850--3856},
  year={2022},
  organization={IEEE}
}

@inproceedings{yasar2020real,
  title={Real-time context-aware detection of unsafe events in robot-assisted surgery},
  author={Yasar, Mohammad Samin and Alemzadeh, Homa},
  booktitle={2020 50th Annual IEEE/IFIP International Conference on Dependable Systems and Networks (DSN)},
  pages={385--397},
  year={2020},
  organization={IEEE}
}

@article{alayrac2022flamingo,
  title={Flamingo: a visual language model for few-shot learning},
  author={Alayrac, Jean-Baptiste and Donahue, Jeff and Luc, Pauline and Miech, Antoine and Barr, Iain and Hasson, Yana and Lenc, Karel and Mensch, Arthur and Millican, Katherine and Reynolds, Malcolm and others},
  journal={Advances in neural information processing systems},
  volume={35},
  pages={23716--23736},
  year={2022}
}

@article{wang2024surgical,
  title={Surgical-LVLM: Learning to Adapt Large Vision-Language Model for Grounded Visual Question Answering in Robotic Surgery},
  author={Wang, Guankun and Bai, Long and Nah, Wan Jun and Wang, Jie and Zhang, Zhaoxi and Chen, Zhen and Wu, Jinlin and Islam, Mobarakol and Liu, Hongbin and Ren, Hongliang},
  journal={arXiv preprint arXiv:2405.10948},
  year={2024}
}

@inproceedings{wang2020towards,
  title={Towards accurate and interpretable surgical skill assessment: A video-based method incorporating recognized surgical gestures and skill levels},
  author={Wang, Tianyu and Wang, Yijie and Li, Mian},
  booktitle={Medical Image Computing and Computer Assisted Intervention--MICCAI 2020: 23rd International Conference, Lima, Peru, October 4--8, 2020, Proceedings, Part III 23},
  pages={668--678},
  year={2020},
  organization={Springer}
}

@article{zhang2021sd,
  title={SD-Net: joint surgical gesture recognition and skill assessment},
  author={Zhang, Jinglu and Nie, Yinyu and Lyu, Yao and Yang, Xiaosong and Chang, Jian and Zhang, Jian Jun},
  journal={International Journal of Computer Assisted Radiology and Surgery},
  volume={16},
  pages={1675--1682},
  year={2021},
  publisher={Springer}
}

@inproceedings{srivastavprocedure,
  title={Procedure-Aware Surgical Video-language Pretraining with Hierarchical Knowledge Augmentation},
  author={Srivastav, Vinkle and Navab, Nassir and Padoy, Nicolas and others},
  booktitle={The Thirty-eighth Annual Conference on Neural Information Processing Systems}
}

@inproceedings{li2023robotic,
  title={Robotic Scene Segmentation with Memory Network for Runtime Surgical Context Inference},
  author={Li, Zongyu and Reyes, Ian and Alemzadeh, Homa},
  booktitle={2023 IEEE/RSJ International Conference on Intelligent Robots and Systems (IROS)},
  pages={6601--6607},
  year={2023},
  organization={IEEE}
}

@article{psychogyios2023sar,
  title={Sar-rarp50: Segmentation of surgical instrumentation and action recognition on robot-assisted radical prostatectomy challenge},
  author={Psychogyios, Dimitrios and Colleoni, Emanuele and Van Amsterdam, Beatrice and Li, Chih-Yang and Huang, Shu-Yu and Li, Yuchong and Jia, Fucang and Zou, Baosheng and Wang, Guotai and Liu, Yang and others},
  journal={arXiv preprint arXiv:2401.00496},
  year={2023}
}

@article{bonrath2013defining,
  title={Defining technical errors in laparoscopic surgery: a systematic review},
  author={Bonrath, Esther M and Dedy, Nicolas J and Zevin, Boris and Grantcharov, Teodor P},
  journal={Surgical endoscopy},
  volume={27},
  pages={2678--2691},
  year={2013},
  publisher={Springer}
}

@article{funke2019video,
  title={Video-based surgical skill assessment using 3D convolutional neural networks},
  author={Funke, Isabel and Mees, S{\"o}ren Torge and Weitz, J{\"u}rgen and Speidel, Stefanie},
  journal={International journal of computer assisted radiology and surgery},
  volume={14},
  pages={1217--1225},
  year={2019},
  publisher={Springer}
}

@article{nwoye2023cholectriplet2021,
  title={CholecTriplet2021: A benchmark challenge for surgical action triplet recognition},
  author={Nwoye, Chinedu Innocent and Alapatt, Deepak and Yu, Tong and Vardazaryan, Armine and Xia, Fangfang and Zhao, Zixuan and Xia, Tong and Jia, Fucang and Yang, Yuxuan and Wang, Hao and others},
  journal={Medical Image Analysis},
  volume={86},
  pages={102803},
  year={2023},
  publisher={Elsevier}
}

@article{loshchilov2017decoupled,
  title={Decoupled weight decay regularization},
  author={Loshchilov, Ilya and Hutter, Frank},
  journal={arXiv preprint arXiv:1711.05101},
  year={2017}
}

@inproceedings{tcn,
  title={Temporal convolutional networks: A unified approach to action segmentation},
  author={Lea, Colin and Vidal, Rene and Reiter, Austin and Hager, Gregory D},
  booktitle={European conference on computer vision},
  pages={47--54},
  year={2016},
  organization={Springer}
}

@article{cares,
  title={Cares: Collaborative agentic reasoning for error detection in surgery},
  author={Low, Chang Han and Zhuo, Zhu and Wang, Ziyue and Xu, Jialang and Liu, Haofeng and Sirajudeen, Nazir and Boal, Matthew and Edwards, Philip J and Stoyanov, Danail and Francis, Nader and others},
  journal={arXiv preprint arXiv:2508.08764},
  year={2025}
}

@inproceedings{resnet,
  title={Deep residual learning for image recognition},
  author={He, Kaiming and Zhang, Xiangyu and Ren, Shaoqing and Sun, Jian},
  booktitle={Proceedings of the IEEE conference on computer vision and pattern recognition},
  pages={770--778},
  year={2016}
}

@inproceedings{clip,
  title={Learning transferable visual models from natural language supervision},
  author={Radford, Alec and Kim, Jong Wook and Hallacy, Chris and Ramesh, Aditya and Goh, Gabriel and Agarwal, Sandhini and Sastry, Girish and Askell, Amanda and Mishkin, Pamela and Clark, Jack and others},
  booktitle={International conference on machine learning},
  pages={8748--8763},
  year={2021},
  organization={PmLR}
}

@article{vaswani2017attention,
  title={Attention is all you need},
  author={Vaswani, Ashish and Shazeer, Noam and Parmar, Niki and Uszkoreit, Jakob and Jones, Llion and Gomez, Aidan N and Kaiser, {\L}ukasz and Polosukhin, Illia},
  journal={Advances in neural information processing systems},
  volume={30},
  year={2017}
}

@article{imitatecholec,
  title={ImitateCholec: A Multimodal Dataset for Long-Horizon Imitation Learning in Robotic Cholecystectomy},
  author={Hansen, Pascal and Kim, Ji Woong Brian and Goldenberg, Antony and Chen, Juo Tung and Li, Yuanzhe Amos and Deguet, Anton and White, Brandon and Tsai, De Ru and Cha, Richard and Jopling, Jeffrey and others},
  journal={Scientific Data},
  year={2026},
  publisher={Nature Publishing Group UK London}
}

@misc{midas,
      title={MiDAS: A Multimodal Data Acquisition System and Dataset for Robot-Assisted Minimally Invasive Surgery}, 
      author={Keshara Weerasinghe and Seyed Hamid Reza Roodabeh and Andrew Hawkins and Zhaomeng Zhang and Zachary Schrader and Homa Alemzadeh},
      year={2026},
      eprint={2602.12407},
      archivePrefix={arXiv},
      primaryClass={cs.RO},
      url={https://arxiv.org/abs/2602.12407}, 
}

\end{document}